\documentclass[10pt,twocolumn,letterpaper]{article}

\usepackage{iccv,times,graphicx,tabulary,multirow,subfig,xspace,xcolor}
\usepackage{amsthm,amssymb,fixmath,mathtools,nicefrac}
\usepackage{placeins}
\usepackage{caption}
\usepackage{epsfig}
\usepackage{booktabs}
\usepackage{blindtext}
\usepackage{makecell}
\usepackage{colortbl}
\usepackage{overpic}
\usepackage{bm}
\usepackage{array}
\usepackage{tabulary}
\usepackage{microtype}
\usepackage[pagebackref=true,breaklinks=true,letterpaper=true,colorlinks,
  citecolor=citecolor,bookmarks=false]{hyperref}
\definecolor{citecolor}{RGB}{34,139,34}

\begin{document}
\title{\\Quality-Aware Network for Face Parsing}
\track{Short-video Face Parsing}

\author{Lu Yang$^1$, Qing Song$^1$, Xueshi Xin$^1$, Wenhe Jia$^1$ and Zhiwei Liu$^2$\\
$^1$Beijing University of Posts and Telecommunications, \\$^2$Institute of Automation Chinese Academy of Sciences.\\
{\tt\small \{soeaver, priv, xinxueshi, jiawh\}@bupt.edu.cn, zhiwei.liu@nlpr.ia.ac.cn}
}

\maketitle

\begin{abstract}
This is a very short technical report, which introduces the solution of the Team BUPT-CASIA for Short-video Face Parsing Track of The 3rd Person in Context (PIC) Workshop and Challenge at CVPR 2021.

Face parsing has recently attracted increasing interest due to its numerous application potentials. Generally speaking, it has a lot in common with human parsing, such as task setting, data characteristics, number of categories and so on. Therefore, this work applies state-of-the-art human parsing method to face parsing task to explore the similarities and differences between them. Our submission achieves 86.84\% score and wins the 2nd place in the challenge.
\end{abstract}

\section{Introduction}
Similar to human parsing~\cite{Yang_cvpr2019_parsingrcnn, Wang_iccv2019_cnif, Wang_cvpr2020_hhp, Yang_eccv2020_rprcnn, Yang_arxiv2021_qanet}, face parsing~\cite{Lin_cvpr2019_facaparsing, Liu_aaai2020_lapa} needs to predict the semantic category of each pixel on the human face, such as mouth, nose, ear, \etc. Short-video Face Parsing\footnote{\fontsize{7pt}{1em}\url{https://maadaa.ai/cvpr2021-short-video-face\\-parsing-challenge/}} (denoted as SFP) is a new face parsing benchmark, which consists of video images collected on the internet, in the format of PNG result images and JSON files. There is a total of 1500 videos, each video has 20 images (1 frame per second for each video). SFP has 19,535 images for training, 2,653 images for validation and 2,525 images for testing with 18 categories (including background). Table~\ref{tab:benchmarks} shows the comparison of SFP with other face parsing benchmarks. 

This work briefly describes the application of state-of-the-art human parsing method to face parsing task, hoping to explore the similarities and differences between them. In addition, our submission achieves 86.84\% score and wins the 2nd place in Short-video Face Parsing Track of The 3rd Person in Context (PIC) Workshop and Challenge at CVPR 2021.

\begin{table}[t]
\centering
\small
\scalebox{0.9}{
\begin{tabular}{c|cc|c}
   Benchmarks       &    Face Num. &  Category Num. &  Sequence ? \\
 \toprule[0.1em]
SFP         & 24,713  & 18 & \checkmark \\
LaPa~\cite{Liu_aaai2020_lapa}       & 22,176 & 11 &  \\
CelebAMask-HQ~\cite{Lee_cvpr2020_maskgan}    & 30,000 & 19 &  \\
\end{tabular}
}
\vspace{.5em}
  \caption{Comparison of SFP with other face parsing benchmarks.}
  \label{tab:benchmarks}
\vspace{-.8em}
\end{table}

\section{Our Approach}

\begin{figure*}
\begin{center}
\includegraphics[width=0.98\linewidth]{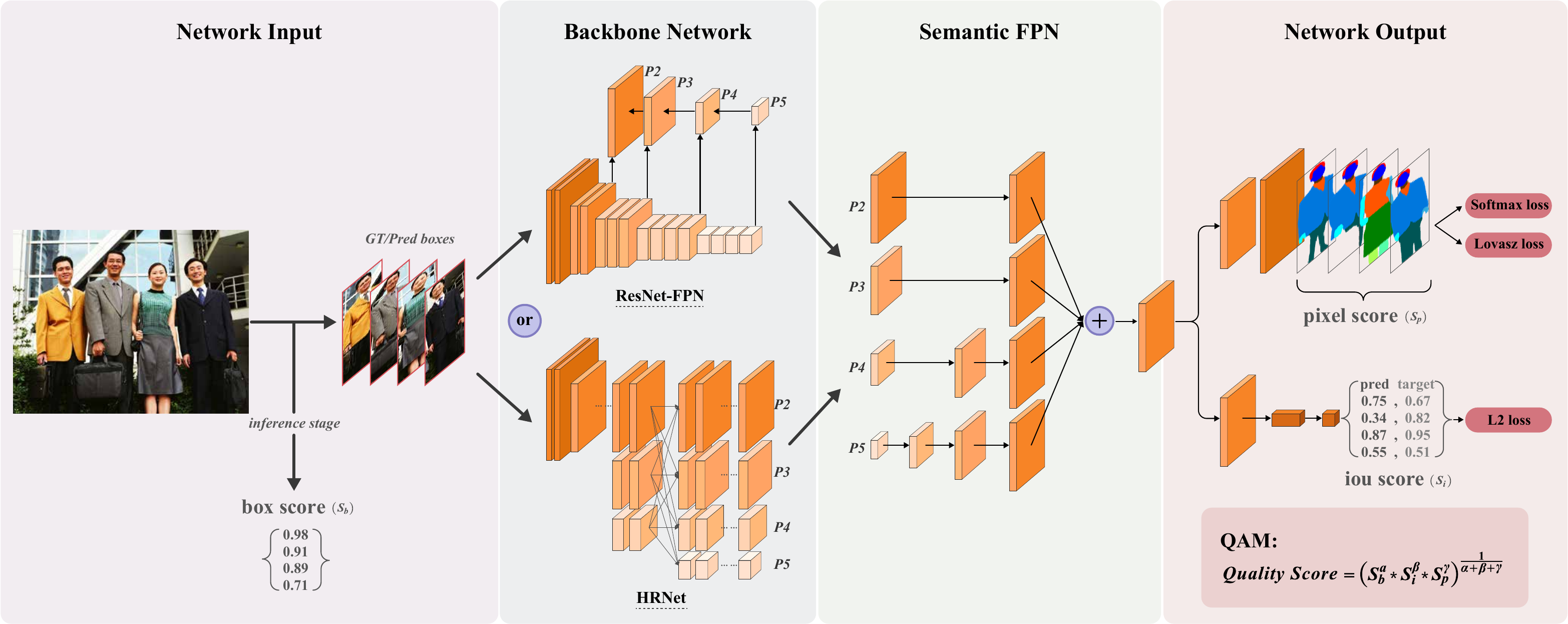}
\end{center}
\vspace{-1mm}
\caption{\textbf{QANet pipeline}. The backbone network outputs multi-scale features through ResNet-FPN or HRNet. Semantic FPN integrates multi-scale features into high-resolution feature, and then predicts parsing result and its IoU score by two branches.}
\label{fig:qanet}
\vspace{-.8em}
\end{figure*}

Due to the high similarity between face parsing and human parsing in task setting and data characteristics, the focus of this work is to study the performance of state-of-the-art human parsing method on face parsing task. QANet~\cite{Yang_arxiv2021_qanet}(\url{https://github.com/soeaver/QANet}) is simple and intuitive in design, and obtains state-of-the-art on several human parsing benchmarks~\cite{Liang_tpami2018_lip, Gong_eccv2018_pgn,  Zhao_mm2018_mhpv2, Xia_cvpr2017_ppp}, so it is selected as the core method of this work. We follow the standard pipeline of QANet (see Figure~\ref{fig:qanet}), and replace the body images with the face images. Please read the original text of QANet for more details. 

 \section{Experiments}
 
 \subsection{Implementation Details}
 
\vspace{3pt}
\noindent\textbf{Dataset.} Due to time constraints, we only use the SFP dataset for experiments.

\vspace{3pt}
\noindent\textbf{Evaluation metric.} The SFP challenge adopts DAVIS J / F score for evaluating the model performance.

\vspace{3pt}
\noindent\textbf{Training.} We conduct experiments based on Pytorch on a server with 8 NVIDIA Titan RTX GPUs. We use HRNet-W48~\cite{Sun_cvpr2019_hrnet} as the backbone, due to its success in dense prediction tasks. Following ~\cite{Xiao_eccv2018_simple}, the ground truth human box is made to a fixed aspect ratio (4 : 3) by extending the box in height or width. The default input size of QANet is 512$\times$384. Data augmentation includes scale [-30\%, +30\%], rotation [-40, +40] and horizontal flip. For optimization, we use the ADAM~\cite{Diederik_iclr2015_adam} solver with 256 batch size. All models are trained for 140 epochs with 2e-3 base learning rate, which drops to 2e-4 at 90 epochs and 2e-5 at 120 epochs. No other training techniques are used, such as warm-up~\cite{Goyal_arxiv2017_1hour}, syncBN~\cite{Chao_cvpr2018_megdet}, learning rate annealing~\cite{Chen_tpami2016_deeplab, Zhao_cvpr2017_pspnet}.

\vspace{3pt}
\noindent\textbf{Testing.} Since most of the faces in the SFP dataset only account for a small part of the image, we choose to use a FCOS~\cite{Tian_iccv2019_fcos} with ResNet50 as detector to detect the face frame first, and then cut it out and send it to QANet. The QANet result is the combination of the original input and the flipped image.

 \subsection{Results}
 
Table~\ref{tab:results} shows the results of Short-video Face Parsing Track. Team TCParser achieves 86.95\% score on the SFP test set and wins the 1st place. Our submission based on QANet is only 0.11 points behind the 1st place and wins the 2nd place in the challenge. The results show that it is effective and successful to directly apply the advanced human parsing method to face parsing task, and also reveal that the two tasks have a high degree of similarity. However, some differences between the two tasks need to be deeply explored, which is also the key to further improve the score.

\begin{table}[t]
\centering
\small
\scalebox{0.98}{
\begin{tabular}{c|c|c}
 Team                       &     Ranking       & Score (\%) \\
 \toprule[0.1em]
TCParser                 & 1st                   & 86.95 \\
BUPT-CASIA (ours) & 2nd                  & 86.84 \\
rat                            & 3rd                   & 86.16 \\
SVFPCC                  &  4th                  & 85.94 \\
hust\_sb                   & 5th                   & 85.87 \\
\end{tabular}
}
\vspace{.5em}
  \caption{Results of Short-video Face Parsing Track, our submission achieves 86.84\% score and wins the 2nd place.}
  \label{tab:results}
\vspace{-.8em}
\end{table}

\begin{table}[t]
\centering
\small
\scalebox{0.98}{
\begin{tabular}{c|ccc|c}
                                                &    Val set      &       Test-aug      &    Ensemble  & Score (\%)   \\
 \toprule[0.1em]
 \multirow{4}{*}{QANet-H48}    &                       &                         &                         & 85.34 \\
                                                 &\checkmark    &                         &                         & 85.66 \\
                                                 &\checkmark    & \checkmark     &                         &  86.08 \\
                                                 &\checkmark    & \checkmark     & \checkmark      & \textbf{86.84}  \\                                                
\end{tabular}
}
\vspace{.5em}
  \caption{The effect of some techniques on our method.}
  \label{tab:techniques}
\vspace{-.8em}
\end{table}

Table~\ref{tab:techniques} shows the effect of some techniques on our method. Standard QANet with HRNet-W48 backbone achieves 85.34\% score on SFP test set. Val set added into training process, test augmentation (multi-scale test) and model ensemble (five models) are improved the results by 0.32 points, 0.42 points and 0.76 points respectively. Finally, our submission achieves 86.84\% score.

 \section{Discussion and Future Work}
This work briefly introduces the effect of QANet on the face parsing task, and gives the scheme of Team BUPT-CASIA participating in Short-video Face Parsing Track of The 3rd Person in Context (PIC) Workshop and Challenge at CVPR 2021. This work hopes to promote the research on the similarities and differences between human parsing and face parsing, and pave the way for a unified solution.

{\small
\bibliographystyle{ieee}
\bibliography{egbib}
}

\end{document}